# Improving Generalization in Visual Reasoning via Self-Ensemble

Tien-Huy Nguyen[1,2], Quang-Khai Tran[3], and Anh-Tuan Quang-Hoang[4]

[1] University of Information Technology, Ho Chi Minh City, Vietnam
[2] Vietnam National University, Ho Chi Minh City, Vietnam
22520567@gm.uit.edu.vn
[3] Ho Chi Minh City Open University, Vietnam
milukhai273@gmail.com
[4] LPL Financial, United States
tuan.quang@lplfinancial.com

**Abstract.** The cognitive faculty of visual reasoning necessitates the integration of multimodal perceptual processing and commonsense and external knowledge of the world. In recent years, a plethora of large vision-language models (LVLMs) have been proposed, demonstrating outstanding power and exceptional proficiency in commonsense reasoning across diverse domains and tasks. Nevertheless, training such LVLMs requires a lot of costly resources. Recent approaches, instead of training LVLMs from scratch on various large datasets, focus on exploring ways to take advantage of the capabilities of many different LVLMs, such as ensemble methods. In this work, we propose self-ensemble, a novel method that improves the generalization and visual reasoning of the model without updating any parameters, a training-free method. Our key insight is that we realized that LVLM itself can ensemble without the need for any other LVLMs, which helps to unlock their internal capabilities. Extensive experiments on various benchmarks demonstrate the effectiveness of our method in achieving state-of-the-art (SOTA) performance on Sketchy-VQA, Outside Knowledge VQA, and out-of-distribution VQA tasks.

**Keywords:** Generalization · Visual Reasoning · Ensemble Learning

## 1 Introduction

Artificial intelligence (AI) has made incredible progress in the past ten years, opening great application possibilities in many sectors, including from healthcare to finance, transportation to education. Many computer vision and natural language processing applications have benefited much from the fast growth of deep learning, especially complex neural network designs including convolutional neural networks (CNNs) and transformers [20,33,64]. Among the several creative artificial intelligence tasks, image retrieval [35, 46], object dectection [45, 49] and visual question answering (VQA) [6,17] have become quite important study subjects attracting a lot of scientific interest. At the confluence of computer vision, natural language processing, and deep learning, VQA a task requiring AI systems to evaluate and combine information from both images and natural language has grown to be a key issue [5, 67]. Beyond the confines of purely academic research, VQA is a necessary first step towards creating artificial intelligence systems able to interact with the environment in a more natural and intuitive way, therefore replicating human cognitive capacities [26].VQA combines several approaches from many AI subfields. It requires a strong knowledge of visual models like ResNet [20] and CLIP [47], language processing architectures like BERT [13] and GPT [8], as well as state-of- the-art multimodal learning approaches [41]. Furthermore, VQA requires strong reasoning capacity, the integration of knowledge from many sources, and the formulation of logical responses all of which are significant challenges in building general artificial intelligence [24].

[0]This research is supported by AI VIETNAM [1].

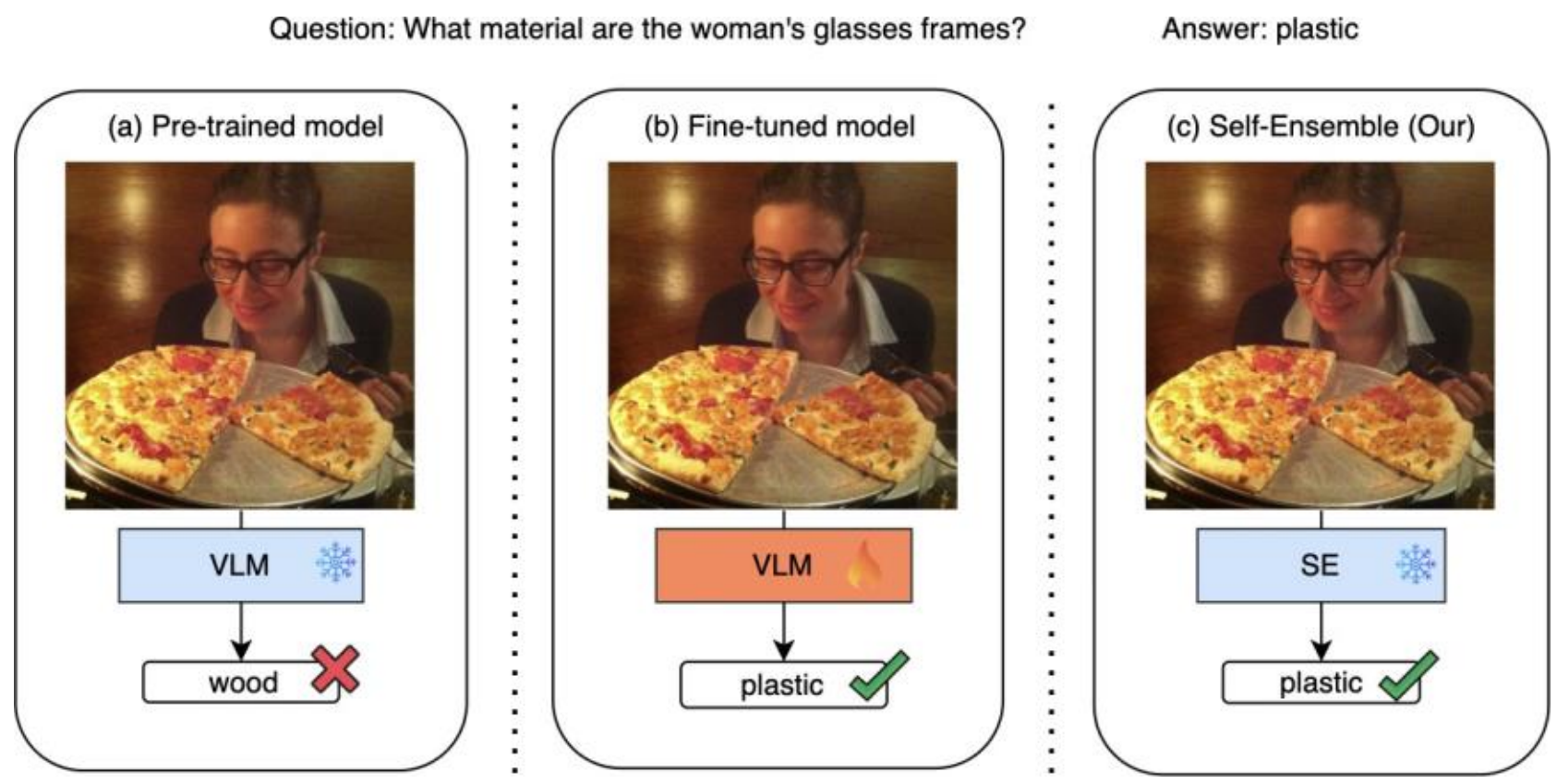

**Fig. 1:** Comparison of different methods in determining the material of eyeglass frames. Part (a) uses a pre-trained VLM model, part (b) is a fine-tuned VLM model, and part (c) is our self-ensemble method. Red Xs indicate incorrect predictions; green ticks indicate incorrect predictions.

VQA has several somewhat broad possible uses. VQA systems could change the way visually impaired people respond to their surroundings in the realm of assistive technology by providing thorough descriptions and responding to questions regarding their surroundings [19]. In terms of healthcare, VQA could help with diagnosis by reviewing medical images and providing specific professional answers to particular problems. Interactive learning systems that let students ask questions about visuals in textbooks and learning materials could be developed in education [29]. Moreover, in e-commerce and visual search, VQA could significantly improve user experience by providing searches depending on intricate image content [16].

Although VQA has made tremendous progress, state-of-the-art technologies still present significant challenges. Above all, most current VQA models require computationally and statistically extensive training. Training these models regularly requires expensive GPU technology and vast manually annotated datasets, therefore restricting their accessibility and wider implementation especially in resource-constrained environments or for highly specialized applications [51].

Second, the inherent complexity and uncertainty of spoken language create a significant obstacle in suitably assessing VQA model performance [53]. Problems with different syntactic patterns but the same semantic meaning could lead to inconsistent responses from VQA models, therefore casting questions on their actual language comprehension capacity [48]. Applying the generalizing capacity of these models to real-world application environments, where language inputs may vary significantly from training data [3]this problem gets more critical.

Finally, present VQA models have a basic challenge in their capacity to answer inquiries outside the training data distribution [21,36]. When presented with novel questions or pictures, several models demonstrate significant performance loss, demonstrating a lack of robust generalization. This underlines the need for solutions that can better adapt to the unexpected and diverse conditions that real-world AI systems constantly confront. To address these challenges, we provide a fresh and innovative paradigm for solving the VQA problem, containing four key contributions:

- A training-free framework that harnesses the strength of existing pre-trained models (as shown in **Figure** 1). This technique helps to considerably decrease computer and data resource requirements.
- We show the utilization of a large language model (LLM) as an intelligent question generator. This unique technique permits the production of many question variations with diverse syntactic structures but related semantic interpretations, leading to a more thorough and trustworthy evaluation of VQA model performance.
- We apply novel self-ensemble approaches to greatly increase the model's potential to answer queries beyond the training data distribution. This method not only enhances accuracy but also boosts the model's flexibility and reliability in varied real-world scenarios.
- To validate the effectiveness of our proposed method, we conducted a suite of comprehensive experiments utilizing both established and novel VQA datasets.

## 2 Related Work

Our work builds upon and contributes to achievements in various critical areas of artificial intelligence, focusing especially on visual reasoning, the generalization capabilities of VQA models, and the effectiveness of ensemble methods in this domain. Below we go deeper into each of these topics, highlighting relevant earlier work and connecting them to our specific contributions.

**Visual reasoning**: Visual reasoning in AI covers the ability of systems to not only see but also comprehend and reason about visual data [43, 75]. Significant gains have been made in core tasks like image classification [57], object detection [38] and scene understanding [78]. These advances serve as the groundwork for handling more complicated visual reasoning difficulties, such as Visual Question Answering (VQA). VQA expands beyond basic object recognition and scene comprehension, requiring models to understand complicated relationships within visual input. To execute visual reasoning with efficacy, a model should integrate the faculties of visual perception and potent deductive reasoning capacities. Traditional visual reasoning models generally depend on intricate architectures [30,71] or demonstrate difficulty in extrapolating from training data. However, contemporary progress in large-scale pre-trained models has demonstrated that vision-language models (VLMs) can attain notable success in visual reasoning tasks even in zero-shot learning scenarios [15, 34, 60]. Our method coincides with continuing efforts to push the frontiers of visual reasoning in VQA, concentrating on problems that demand a deep comprehension of spatial relationships, inferring object functionality, and reasoning about dynamic events and activities, as reported in recent research [50, 55, 72].

**VQA Generalization**: A continuing problem in VQA research is strengthening the models' capacity to generalize effectively to novel questions and visual environments [4, 31, 54, 63]. While models perform well on training data, they sometimes struggle with new queries or photos that dramatically diverge from the training distribution. This fragility underscores the necessity for measures aimed at enhancing the resilience and generalization capabilities of VQA models. Recent research has examined different approaches, including data augmentation strategies to boost training data diversity [11, 79], compositional model architectures to decompose complicated reasoning problems [27], and knowledge-based methods to enrich models with external information [9, 65]. Our research contributes to this crucial area by investigating the possibility of ensemble techniques to boost VQA model generalization. By intelligently mixing predictions from different models, we seek to construct a more resilient system capable of handling a broader spectrum of queries and scenarios, inspired by successful ensemble learning implementations [80].

**Ensemble**: Ensemble learning has continually proved its potential across multiple machine learning problems, typically exceeding individual models in terms of performance and resilience. The primary premise of ensemble methods is to aggregate predictions from several, often varied models, capitalizing on their particular strengths and reducing their faults. This method has been successful in image classification [68, 74], where ensembles of Convolutional Neural Networks have attained state-of-the-art results, and in natural language processing [22], where ensembles of language models have enhanced tasks such as sentiment analysis and machine translation. Inspired by these accomplishments, we study the applicability of ensemble approaches inside visual reasoning and VQA. Our study focuses on establishing effective ways for integrating model predictions, with the goal of improving both the accuracy and generality of VQA systems.

# 3 Proposed Method

## 3.1 Optimal Question Formulation in VQA

In the domain of **Visual Question Answering** (VQA), question formulation has an important role in determining the model's accuracy. The ability to generate diverse semantically equivalent questions will enhance the robustness of the VQA system by accepting multiple variations of natural language inputs. The diversity of language will impact people's perceptions about formulating questions. Different cultures or different countries will have various ways to ask questions. That difference will lead to many ways of asking questions about topics and answering. For instance, in **Figure** 2, there are two questions but their meaning are identical. The purpose of this question is to ask about the color of the vehicle on the lot. However, the traditional VLM will shift questions into two distinct answers (black or green). The correct answer should be black according to the image. However, because of the different ways of asking will change the correct answer. Our question is **"Which question is optimal and how can we formulate an optimal question?**

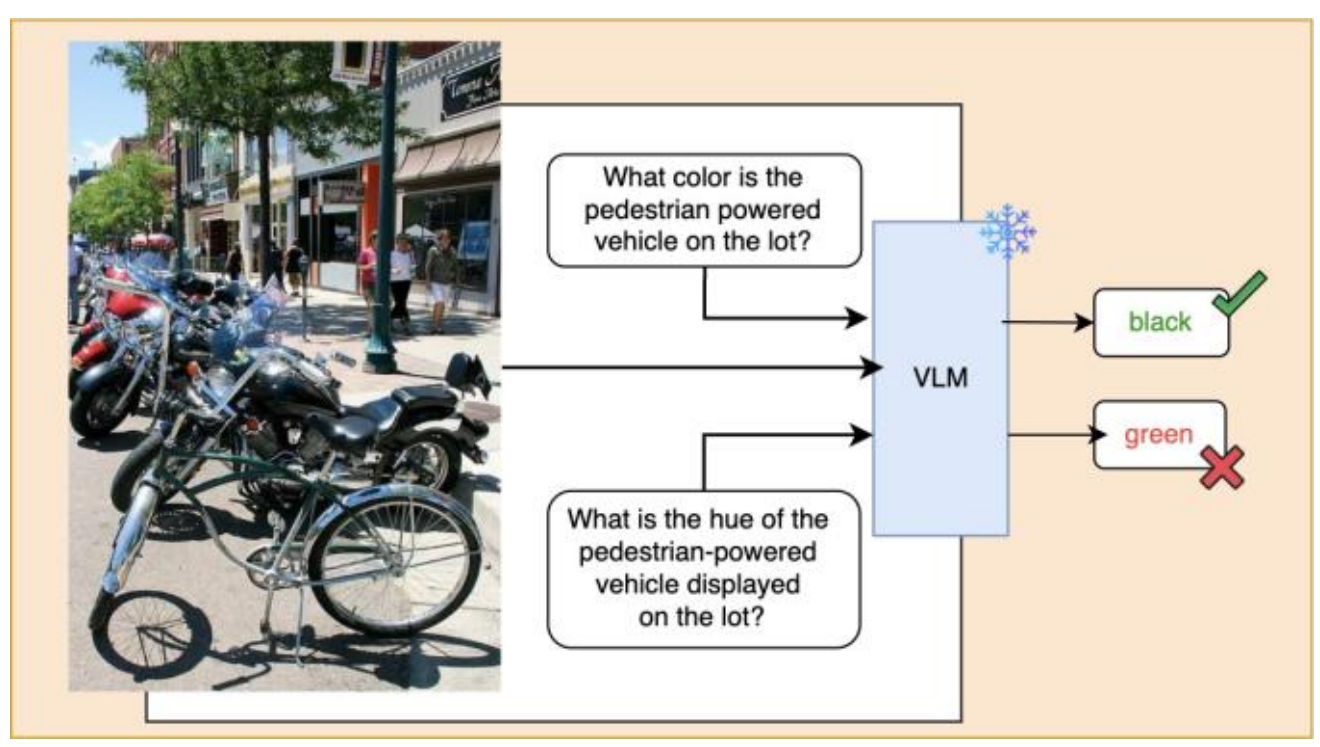

**Fig. 2:** Illustration of polysemy in language for the VLM model. Two questions about the same object (bicycle) were worded differently, resulting in different answers from the model. The question "What color is the pedestrian powered vehicles on the ground?" received the correct answer "black", while the question "What color is the hue of the pedestrian-powered vehicle displayed on the lot?" received the incorrect answer "green". This shows that the way the question is worded can influence the results from the AI model.

## 3.2 Question Generator

**In-context learning** (ICL) is a fascinating capability of large language models (LLMs) to learn new tasks without any parameter updates. This means that instead of traditional retraining on the new tasks, the model is fed a few examples of the desired task directly within the input prompt and gleans knowledge from the examples provided in the prompt itself. This allows for remarkable flexibility and adaptation, enabling LLMs to perform tasks they haven't been explicitly trained for, simply by understanding the patterns and relationships demonstrated in the given examples. Therefore, we take advantage of the impressive capabilities of LLMs, treating them as a **Question Generator (QG)**, to generate other questions.

| Prompt template |
| --- |
| prompt = exemplar + "Please rewrite the above question into N other questions but keep the same semantics." |

**Table 1:** illustrates how to generate multiple variations of a question while retaining the original meaning, enhancing the model's natural language processing capabilities

| Initial Question | Generated Question |
| --- | --- |
| 0. Where are the people laying? | 1. What is the location where the people are resting?<br>2. Where are the people stretched out? |
| 0. How many bicycles are there in the image? | 1. How many bicycles are evident in the photograph?<br>2. What's the count of bicycles that appear in the image? |
| 0. What century were these were invented in? | 1. In what period were these first created?<br>2. When did the invention of these begin? |

**Table 2:** The left column is the original question from the dataset, and the right column is the new question generated by the Question Generator. Each original question is converted into two new questions, keeping the same meaning but with different wording. This shows the system's ability to generate semantically compatible questionable variables.

**Prompt template** is described in **Table 1**. First, we simply add a few examples to take advantage of LLM's adaptability in new tasks, by applying in-context learning capabilities. We then designed an instruction prompt for QG to understand the requirements of the new task and generate appropriate questions.

As we analyzed in **Section 3.1**, a different way of asking questions, for each question will lead to a different answer, in order to respond to correct answers, it is necessary to formulate optimal questions. Therefore, we hypothesize that, compared to the original question, other questions that retain the same semantics represent a different observational perspective, which leads to optimal answers.

Therefore, we designed such instructional prompts to generate additional questions but still retain the same semantic features as the initial question.

$$q = QG(prompt(q_0, e, p)) \tag{1}$$

The question generation process is described in **Equation 1**. For each initial question $q_0$, along with a prompt designed like a prompt template receives a few examples **e** and instruction prompt **p** so that QG generates N questions.

As shown in **Table 2**, the Question Generator maintains the core meaning by Lexical Substitution, Syntactic Variation, and Semantic Preservation while paraphrasing the initial questions. Overall, the generated questions are effective in testing the same conceptual understanding as the original questions while providing different perspectives and contexts. This variety can help in improving the robustness and flexibility of a model's ability to answer questions related to the same topic.

### 3.3 Self-Ensemble

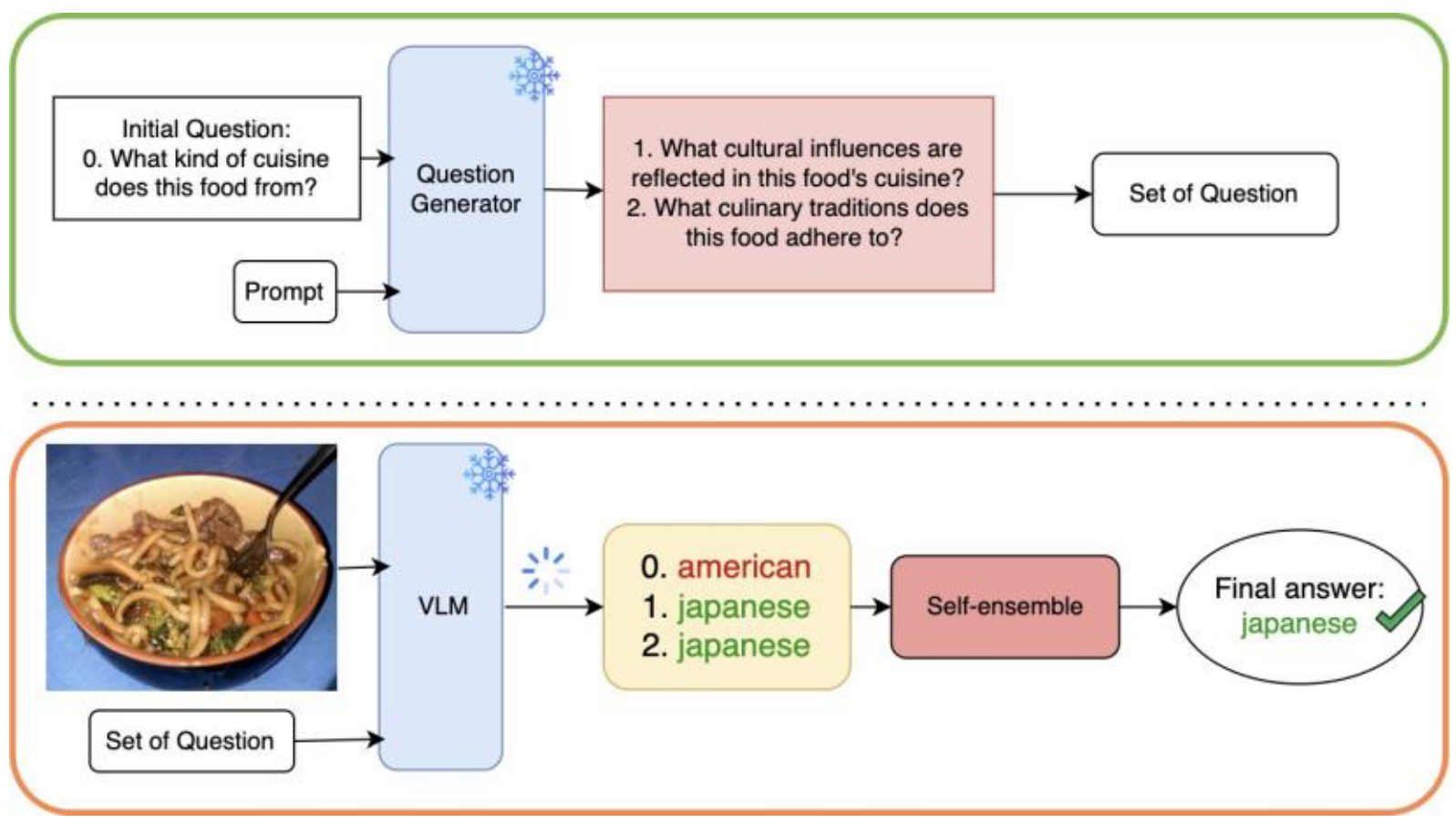

**Fig. 3:** Overview of our system, including 2 stages. The top part of the image is stage 1 described in **Section** 3.2 and the bottom part of the image is stage 2 described in **Section** 3.3. The red text corresponds to a false answer, the green text displayed is the correct answer. The number 0 indicates that it was the initial question/answer. Numbers 1 and 2 refer to questions/answers generated by Question Generator and VLM.

**Overall System** We use Gemma [59] 7B parameters model from Google for Question Generator and QwenChat [7] from Alibaba for Vision Language Model (VLM). For both models, we use pretrained model. Before generating the answer, we create a set of questions **Q** with the same semantic features as the initial question as depicted in **Figure** 3. In stage 2, self-ensemble is used to produce the final answer.

**Vision Language Model** will provide plausible answers to the questions received from stage 1. The process of generating answers to questions is described in **Equation** 2.

$$a = VLM(i, q) \tag{2}$$

In which, for each **q** in the set of questions **Q** and the corresponding image **i**, each answer **a** corresponding to each question will be returned in turn. Finally, we obtain an answer set of answers **A**.

Ensemble voting is a technique in ensemble learning where the final prediction is made by combining the predictions of multiple individual models, where majority voting will choose the prediction with the most predictions of the models as the final prediction. While **Self-ensemble** will return one final result based on a majority voting mechanism but uses only one Vision Language Model (VLM). This shows that VLM can self-ensemble without using many other models.

**Algorithm 1** Self-ensemble for selecting the final answer based on majority voting mechanism.

```
require: set of answer A, examplars E
ensure: one final answer
prompt = E + "let give me the number of answers above that have the most
    similar meaning and subset of those answers."        ▷ prompt initialization
SE = large language model                                ▷ self-ensemble initialization
num, subset = SE(A, prompt)
if num is equal to 1                                     ▷ random selection
    return one of the answers in A
else if num is larger than 1                             ▷ majority voting
    return one of the answers in subset
```

**The self-ensemble** algorithm is described in **Algorithm 1**. We design self-ensemble as essentially a Large Language Model (LLM). To help LLM better understand the task requirements, we use their in-context learning capabilities by initializing a prompt with examples. Self-ensemble will then receive a set of answers A and the instruction prompt as input and provide the number of answers with the same semantic meaning along with the subset of answers. This means that the subset of answers is the majority in the set of answers A. But if that subset only includes one answer, that is equivalent to $num = 1$, corresponding to not having a majority answer set, we randomly select one of the answers in the set of answers A.

We designed to ask and answer many times for VLM before giving the final answer, intending to help VLM itself observe many times, with many different views, and above all, help VLM to think many times, help improve visual reasoning, generalization, and help stabilize consistency level.

As depicted in **Figure 3**, VLM initially answered incorrectly when asked about the origin of the food, but after applying self-ensemble, VLM thought many times about the same problem with different questions, generated a set of answers A, and finally coming up with the correct answer 'Japanese', based on majority voting, choose the Japanese answer because it accounts for the majority of the answer set.

## 4 Experimental Results

### 4.1 Experimental Setup

To conduct our experiments and evaluate the performance of our self-ensemble method, we utilized a high-performance computing setup designed to efficiently handle visual data processing tasks with the dataset. Our experiments were conducted on a machine running Ubuntu 20.04, equipped with an NVIDIA RTX 4090 GPU featuring 24 GB of VRAM and 128 GB of RAM. This robust hardware configuration was essential to accommodate the computational demands of the baseline model, particularly when processing large datasets and executing intensive training and inference tasks.

We experiment on 5 different datasets. **OODCV-VQA**: This dataset focuses on out-of-distribution visual question answering, assessing the model's ability to process unfamiliar images and questions. It encompasses images and queries from less common domains or contexts, challenging the model's generalization capacity [62]. **OODCV-Counterfactual**: Extending the OOD concept, this dataset concentrates on counterfactual scenarios in computer vision contexts. It comprises image-question pairs that require the model to reason about hypothetical situations or alterations within images, evaluating its complex inference capabilities [62]. **Sketchy-VQA**: This unique dataset employs sketches or simple drawings instead of real images. It challenges the model's ability to understand and answer questions based on abstract or incomplete information, demanding a high level of abstraction capability [62]. **Sketchy-challenging**: An advanced iteration of Sketchy-VQA, this dataset incorporates more complex questions and sketches. It demands a higher level of reasoning and deeper understanding from AI models, challenging their ability to process abstract information at an elevated level [62]. **A-OKVQA**: This dataset combines visual question answering with open-knowledge requirements. It not only demands that models understand image content but also integrate general knowledge to answer questions. A-OKVQA includes queries related to culture, history, science, and other fields, extending beyond the direct information present in the images [52].

### 4.2 Quantitative Results

In **Table 3**, we first observe that our method achieves State-of-the-art (SOTA) performance on the two datasets OODCV-VQA and Sketchy Challenge by only applying self-ensemble, without needing to train and update any parameters. We use the pertained Qwen model as a Vision Language Model (VLM). On the contrary, other models to achieve SOTA performance use baseline models with a large number of parameters (v1-Vicuna-vo-13B, Vicuna- v.15-13B, LLaMA-Chat-13B).

| **Method** | **Language Model** | **Params** | **Vision Model** | **Params** | **OOD-CV test** | **OOD-CV Counterfact test** | **Sketchy test** | **Sketchy Challenge test** |
|---|---|---|---|---|---|---|---|---|
| Vilt [32] | BERT [13] | 110M | ViT [14] | 632M | 3,97 | 18,43 | 49,9 | 50,01 |
| v1-Vicuna-v0-13B [40] | MiniGPT4 [81] | 14B | EVA-CLIP [56] | 1.3B | 39,97 | 50,62 | 60,3 | 44,3 |
| v1-Vicuna-v0-7B [40] | MiniGPT4 [81] | 8B | EVA-CLIP [56] | 1.3B | 41,74 | 36,03 | 64,4 | 46,8 |
| v2-LLaMA-Chat-7B [61] | MiniGPT4 [81] | 14B | EVA-CLIP [56] | 1.3B | 52,3 | 36,03 | 66,9 | 63 |
| Fuyu [2] | Fuyu [2] | 8B | Fuyu [2] | 8B | 54,38 | 19,87 | 65,9 | 57,2 |
| mPLUG-Owl [69] | mPLUG-Owl [69] | 8.2B | OpenAI-CLIP [47] | 428M | 54,75 | 45,64 | 70,8 | 60,7 |
| BLIP2 [37] | FLANT5 [66] | 11B | EVA-CLIP [56] | 1.3B | 55,21 | 47,41 | - | - |
| LLaMA-Adapter [77] | LLaMA-Adapter [77] | 7.2B | OpenAI-CLIP [47] | 428M | 55,25 | 42,39 | 77,5 | 69,8 |
| Vicuna-v0-7B [40] | LLaVA [39] | 7.2B | OpenAI-CLIP [47] | 428M | 56,16 | **60,72** | 83,2 | 82,2 |
| FlanT5-XXL [66] | InstructBLIP [12] | 14B | EVA-CLIP [56] | 1.3B | 57,77 | 51,31 | 89,5 | 87,9 |
| v1-LLaMA-Chat-7B [61] | MiniGPT4 [81] | 8B | EVA-CLIP [56] | 1.3B | 57,87 | 44,62 | 72,9 | 70,6 |
| LLaMA-Chat-13B [61] | LLaVA [39] | 13.4B | OpenAI-CLIP [47] | 428M | 63,93 | 40,89 | 89,6 | 86,9 |
| Vicuna-v1.5-7B [40] | LLaVA [39] | 7.2B | OpenAI-CLIP [47] | 428M | 70,26 | 46,62 | 80,4 | 80,3 |
| mPLUG-Owl2 [70] | mPLUG-Owl2 [70] | 8.2B | OpenAI-CLIP [47] | 428M | 71,08 | 41,9 | <u>91,3</u> | <u>87,5</u> |
| FlanT5-XL [66] | InstructBLIP [12] | 12B | EVA-CLIP [56] | 1.3B | 71,44 | 48,07 | 88,5 | 83,9 |
| InternLM-X [76] | InternLM-X [76] | 8B | EVA-CLIP [56] | 1.3B | 71,57 | 43,38 | 84,3 | 75,6 |
| Vicuna-v1.5-13B [40] | LLaVA [39] | 13.4B | OpenAI-CLIP [47] | 428M | 71,79 | 47,7 | **91,5** | 84 |
| Vicuna-v0-7B [40] | InstructBLIP [12] | 8B | EVA-CLIP [56] | 1.3B | 74,92 | 52,69 | 87,2 | 82,2 |
| QwenChat [7] | LLaVA [39] | 7B | OpenAI-CLIP [47] | 428M | <u>76.07</u> | 56.66 | 86.7 | 86.6 |
| **Our** | VLM: QwenChat [7] | 8B | LLM: Gemma [59] | 7B | **79.18** | <u>59.54</u> | 90.4 | **88.9** |

**Table 3:** The table details the methods, including the model name, the type of language and vision model of Vision Language Model (VLM) we used, and the number of parameters. The final columns show the accuracy of datasets such as OOD-CV, Sketchy, and their variants. We set N = 2 and conduct experiments on 4 test sets of datasets. **The bolded number** and <u>the underlined number</u> are the highest score and the second highest score, respectively. The results show significant differences in performance between the models. This provides an overview of how well different methods perform on visual reasoning tasks, allowing for direct comparisons between models on the same benchmark set.

On the OOD-CV Counterfact and Sketchy datasets, it gives relatively competitive results compared to the SOTA model's performance when reaching 59.54% and 90.4%, challenging compared to most other models, when before and after applying, our method achieved better results than Vicuna-v0-7B, FlanT5-XL, Flaa-T5-XXL, and LLaMA-Chat-13B.

When compared to QwenChat, we increase 3.11%, 2.88%, 3.7%, and 2.3% on the OOD-CV, OOD-CV Counterfact, Sketchy, and Sketchy Challenge datasets, respectively.

**Table** 4 contains the comparative results on the challenging A-OKVQA dataset, an augmented version of the OK-VQA dataset, which requires outside knowledge. We compare our method to the strong baselines and the current SOTA methods. The results show the superiority of our method over the counterparts on multiple-choice question tasks, reflecting the effectiveness and generalization of our method.

From the Question Generator, we choose N = 4 to generate other questions with the same semantics, along with the initial question. We have a total of five questions, with the implication that VLM can think carefully five times before giving the final answer by applying self-ensemble.

We achieve SOTA performance compared to previous methods. Previous improvement methods mostly require model training on many epochs(COLA-FT, CLIPcap, ViLBERT, LXMERT, GPV-2..), especially pre-training on Visual

| Method | Accuracy |
|---|---|
| Img2Prompt [18] | 42.9 |
| Pythia [25] | 49.0 |
| ViLBERT [41] | 49.1 |
| LXMERT [58] | 51.4 |
| KRISP [42] | 51.9 |
| CLIPcap [44] | 56.9 |
| Prophet [73] | 59.3 |
| GPV-2 [28] | 60.3 |
| COLA-ZERO [10] | 72.3 |
| Prophet-MC [73] | 76.4 |
| COLA-FT [10] | 78.1 |
| PaLI-X-VPD-5B [23] | 79.7 |
| **Our (N=4)** | **84.11** |

**Table 4:** Accuracy of different methods on A-OKVQA benchmark. We experimented with the results of the validation set and multiple-choice question version of the A-OKVQA dataset.

Question Answering datasets. Based on our method, which is the training-free method, no training is required, reducing the cost of resources used and training time.

Multi-think, multi-observation, and self-ensemble help the model improve visual reasoning, in common cases, as well as cases that are out-of-distribution scenarios in OODCV-VQA, OOD-CV Counterfact, Sketchy, and Sketchy Challenge datasets or need outside knowledge in A-OKVQA Dataset.

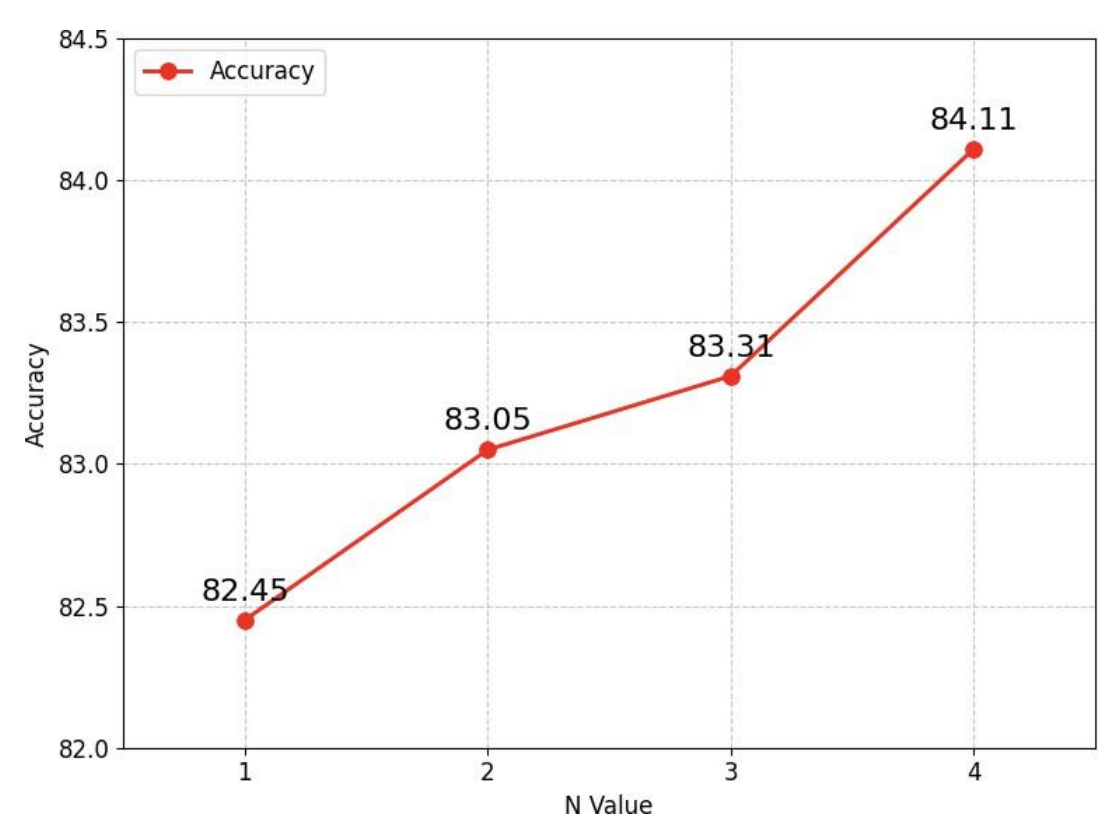

**Fig. 4:** The graph shows the relationship between the N value and accuracy of the model on the validation set of the A-OKVQA dataset. The N value on the horizontal axis shows the number of questions generated by Question Generator and the accuracy value corresponds to the N value on the vertical axis.

We conducted experiments and analyzed the influence of the N value on the performance of the model which is shown in **Figure** 4. Along with the initial question, we collected a set of questions including $N + 1$ questions.

We observe that the model's accuracy will increase sequentially as the N values gradually increase. When N = 1 reaches 82.45% and when N = 4 reaches 84.11% an increase of 1.66%. That means making VLM think more will help improve the model's performance, as well as the effectiveness of our proposed method.

### 4.3 Quanlitative Results

In this section, we experiment with the hypothesis we propose in section 3. By examining specific case studies, we illustrate how our **self-ensemble** method enhances the model's ability to understand and interpret the content based on input questions. In our experiment, we include six examples that represent different scenarios in VQA. To understand the performance and accuracy of the model, we try to divide examples into three case studies.

- Case Study 1: Understanding Complex Scene
  A significant advantage of our approach is its capability to manage complex scenes containing multiple objects and interactions. In Figure 5, we illustrate the model's ability to correctly identify the number of chairs in a scene when an additional chair is hypothetically added. To assess the model's consistency, we generated five different questions, all with the same semantic meaning.
  Although the model initially provided incorrect answers to questions 0 and 3, identifying the count as "four" instead of the correct "five," our self-ensemble method successfully resolved these discrepancies. By applying a voting mechanism, the self-ensemble approach selected "five" as the final answer, since it appeared in three out of the five responses, compared to "four," which appeared in only two. This demonstrates the robustness of our method in choosing the most probable correct answer.
  Additionally, the model's performance on an image of the sky, populated with airplanes, further showcases its ability to comprehend complex visual scenes. The model accurately counted the number of airplanes present, highlighting its enhanced visual reasoning capabilities when faced with intricate spatial arrangements and numerous elements.
- Case Study 2: Handling Ambiguous Queries
  Our self-ensemble method demonstrates exceptional robustness in addressing ambiguous or polysemous questions. For instance, when a user inquires, "What information does the electric street sign display?" as shown in Figure 5, our model consistently provides the correct answer, even when the question is rephrased into more ambiguous forms such as "What message is conveyed by the electric sign?" or "What words or symbols appear on the illuminated street sign?" Despite the potential for different interpretations, the model accurately identifies the correct response as "don't walk," avoiding incorrect alternatives like "proceed."

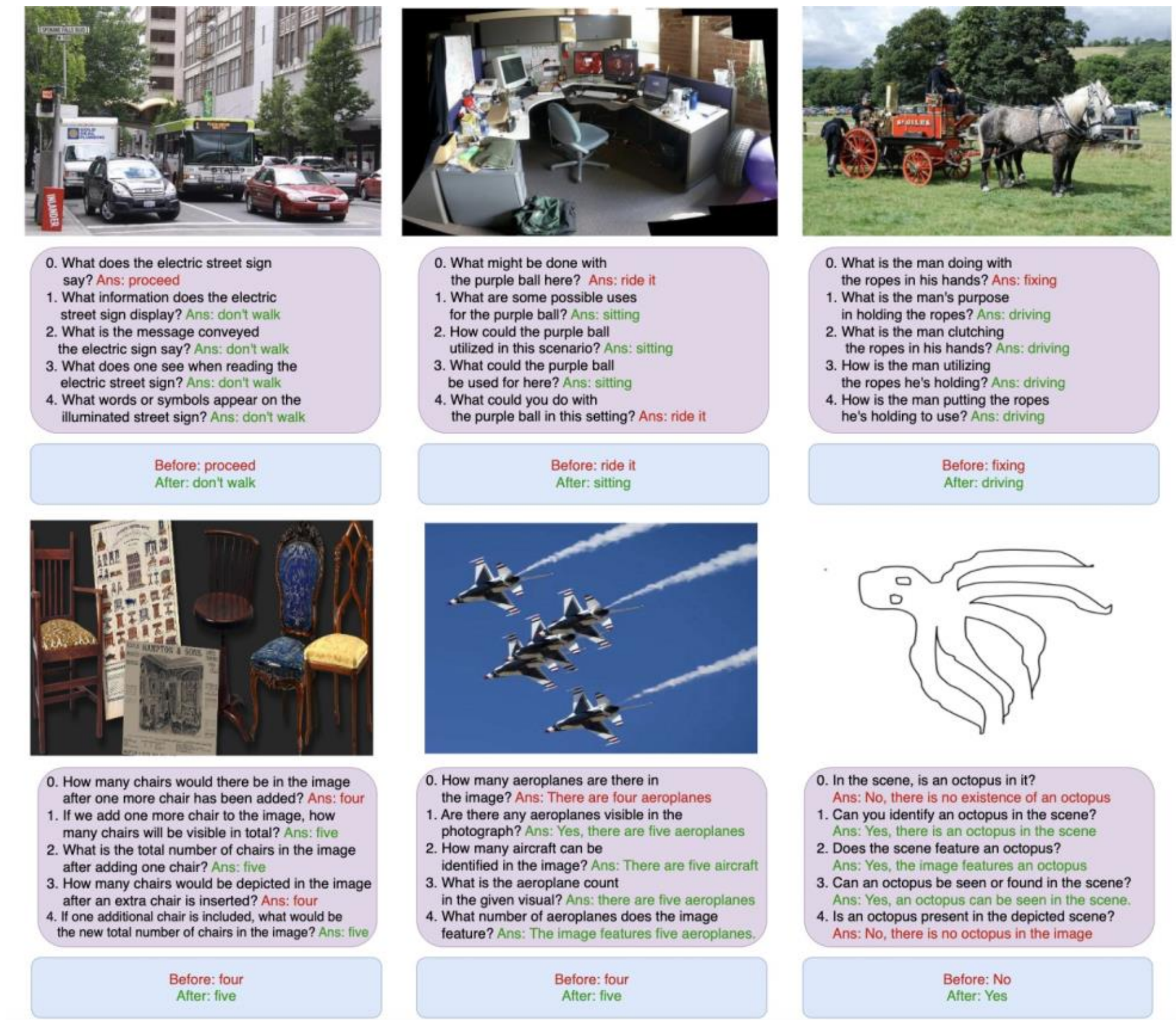

**Fig. 5:** Example of a variable question generation technique to improve model generalization in a VQA task. Consists of six different images, each accompanied by a series of related questions worded in multiple ways. This technique aims to enhance the model's ability to understand and answer questions that have the same meaning but are worded differently. By selecting the answer that appears most frequently in the question variations, the model can learn to produce more consistent and accurate answers, thereby improving its generalization across diverse question-answering scenarios.

Moreover, Figure 5 presents a scenario where the model is asked about the function of a purple ball. Given that the purple ball could serve various purposes, the model effectively utilizes contextual information from the image to determine "sitting" as the appropriate function. This capability highlights the model's proficiency in interpreting context and resolving ambiguity, allowing it to provide accurate responses even in situations with multiple possible interpretations.

These examples demonstrate the effectiveness of our self-ensemble method in dealing with linguistic ambiguity and polysemy, emphasizing its ability to deliver accurate answers regardless of varied question formulations. By leveraging context and semantic understanding, the model showcases advanced interpretative skills that enhance its reliability in real-world applications where ambiguity is common.

- Case Study 3: Interpreting Abstract Sketches
To verify the model's ability to interpret abstract information, we employed the Sketchy-VQA dataset. The model was asked to describe a sketch depicting a simple octopus. It correctly identified the octopus when we increased the number of questions. Although some responses are not correct, the final result still returns a correct answer "octopus", which is a desired answer that we look for.

The qualitative results highlight several key strengths of our self-ensemble VQA approach, including enhanced robustness in handling linguistic variability, and the ability to interpret abstract sketches. These findings demonstrate the potential of our method to overcome some of the limitations faced by existing VQA models, particularly in terms of generalization and adaptability.

However, certain challenges remain, such as accurately interpreting scenes with high visual complexity or nuanced interactions. Future work should focus on further refining the model's ability to handle such challenges, possibly by integrating additional contextual information or employing more sophisticated reasoning mechanisms.

## 5 Dicussion

**Limitation:** Our method takes advantage of the in-context learning capabilities of LLMs. On the contrary, LLMs provide improved performance, increasing the ability to generalize across datasets by using 2-shot prompts in the question generation process, but when implementing a zero-shot setting, there are many limitations.

**Future Work:** First, to increase the generalization ability of the model, explore the most optimal generation and selection of questions in the zero-shot setting. Second, you can expand your thinking ability to other in-depth tasks such as DocumentVQA and CountVQA. Third, by applying the process of generating a reasonable rationale, to help LLMs become more interpretable and generally understood.

**Conclusion:** In this paper, we propose a new "self-ensemble" method for visual reasoning task, VQA, which helps to explore the internal potential of the model using only VLM, without using many other models. In addition, self-ensemble is a training-free method, helping to save computational resources for model training. LLM acts as a reasoner and selects the final answer based on majority voting in self-ensemble. Experiments show that reasoning performance improves on various datasets. The results demonstrate the effectiveness of our proposed method to improve the model's generalization.

## References

1. AI VIETNAM — aivietnam.edu.vn. https://aivietnam.edu.vn 1
2. Fuyu-8B: A Multimodal Architecture for AI Agents — adept.ai. https://www.adept.ai/blog/fuyu-8b, [Accessed 10-08-2024] 10
3. Agrawal, A., Batra, D., Parikh, D., Kembhavi, A.: Don't just assume; look and answer: Overcoming priors for visual question answering (2018), https://arxiv.org/abs/1712.00377 3
4. Akula, A., Changpinyo, S., Gong, B., Sharma, P., Zhu, S.C., Soricut, R.: CrossVQA: Scalably generating benchmarks for systematically testing VQA generalization. In: Moens, M.F., Huang, X., Specia, L., Yih, S.W.t. (eds.) Proceedings of the 2021 Conference on Empirical Methods in Natural Language Processing. pp. 2148–2166. Association for Computational Linguistics, Online and Punta Cana, Dominican Republic (Nov 2021). https://doi.org/10.18653/v1/2021.emnlp-main.164, https://aclanthology.org/2021.emnlp-main.164 4
5. Anderson, P., He, X., Buehler, C., Teney, D., Johnson, M., Gould, S., Zhang, L.: Bottom-up and top-down attention for image captioning and visual question answering. In: 2018 IEEE/CVF Conference on Computer Vision and Pattern Recognition. pp. 6077–6086 (2018). https://doi.org/10.1109/CVPR.2018.00636 2
6. Antol, S., Agrawal, A., Lu, J., Mitchell, M., Batra, D., Zitnick, C.L., Parikh, D.: Vqa: Visual question answering. In: Proceedings of the IEEE International Conference on Computer Vision (ICCV) (December 2015) 2
7. Bai, J., Bai, S., Chu, Y., Cui, Z., Dang, K., Deng, X., Fan, Y., Ge, W., Han, Y., Huang, F., Hui, B., Ji, L., Li, M., Lin, J., Lin, R., Liu, D., Liu, G., Lu, C., Lu, K., Ma, J., Men, R., Ren, X., Ren, X., Tan, C., Tan, S., Tu, J., Wang, P., Wang, S., Wang, W., Wu, S., Xu, B., Xu, J., Yang, A., Yang, H., Yang, J., Yang, S., Yao, Y., Yu, B., Yuan, H., Yuan, Z., Zhang, J., Zhang, X., Zhang, Y., Zhang, Z., Zhou, C., Zhou, J., Zhou, X., Zhu, T.: Qwen technical report (2023), https://arxiv.org/abs/2309.16609 7, 10
8. Brown, T.B., Mann, B., Ryder, N., Subbiah, M., Kaplan, J., Dhariwal, P., Neelakantan, A., Shyam, P., Sastry, G., Askell, A., Agarwal, S., Herbert-Voss, A., Krueger, G., Henighan, T., Child, R., Ramesh, A., Ziegler, D.M., Wu, J., Winter, C., Hesse, C., Chen, M., Sigler, E., Litwin, M., Gray, S., Chess, B., Clark, J., Berner, C., McCandlish, S., Radford, A., Sutskever, I., Amodei, D.: Language models are few-shot learners (2020), https://arxiv.org/abs/2005.14165 2
9. Buck, C., Bulian, J., Ciaramita, M., Gajewski, W., Gesmundo, A., Houlsby, N., Wang, W.: Ask the right questions: Active question reformulation with reinforcement learning (2018), https://arxiv.org/abs/1705.07830 4
10. Chen, L., Li, B., Shen, S., Yang, J., Li, C., Keutzer, K., Darrell, T., Liu, Z.: Large language models are visual reasoning coordinators (2023), https://arxiv.org/abs/2310.15166 11
11. Cubuk, E.D., Zoph, B., Mané, D., Vasudevan, V., Le, Q.V.: Autoaugment: Learning augmentation strategies from data. In: 2019 IEEE/CVF Conference on Computer Vision and Pattern Recognition (CVPR). pp. 113–123 (2019). https://doi.org/10.1109/CVPR.2019.00020 4
12. Dai, W., Li, J., Li, D., Tiong, A.M.H., Zhao, J., Wang, W., Li, B., Fung, P., Hoi, S.: Instructblip: Towards general-purpose vision-language models with instruction tuning (2023), https://arxiv.org/abs/2305.06500 10
13. Devlin, J., Chang, M.W., Lee, K., Toutanova, K.: Bert: Pre-training of deep bidirectional transformers for language understanding (2019), https://arxiv.org/abs/1810.04805 2, 10

14. Dosovitskiy, A., Beyer, L., Kolesnikov, A., Weissenborn, D., Zhai, X., Unterthiner, T., Dehghani, M., Minderer, M., Heigold, G., Gelly, S., Uszkoreit, J., Houlsby, N.: An image is worth 16x16 words: Transformers for image recognition at scale (2021), https://arxiv.org/abs/2010.11929 10
15. Du, Y., Li, J., Tang, T., Zhao, W.X., Wen, J.R.: Zero-shot visual question answering with language model feedback (2023), https://arxiv.org/abs/2305.17006 4
16. Gao, Y., Beijbom, O., Zhang, N., Darrell, T.: Compact bilinear pooling. In: 2016 IEEE Conference on Computer Vision and Pattern Recognition (CVPR). pp. 317–326 (2016). https://doi.org/10.1109/CVPR.2016.41 2
17. Goyal, Y., Khot, T., Summers-Stay, D., Batra, D., Parikh, D.: Making the v in vqa matter: Elevating the role of image understanding in visual question answering. In: 2017 IEEE Conference on Computer Vision and Pattern Recognition (CVPR). pp. 6325–6334 (2017). https://doi.org/10.1109/CVPR.2017.670 2
18. Guo, J., Li, J., Li, D., Tiong, A.M.H., Li, B., Tao, D., Hoi, S.C.H.: From images to textual prompts: Zero-shot vqa with frozen large language models (2023), https://arxiv.org/abs/2212.10846 11
19. Gurari, D., Li, Q., Stangl, A.J., Guo, A., Lin, C., Grauman, K., Luo, J., Bigham, J.P.: Vizwiz grand challenge: Answering visual questions from blind people (2018), https://arxiv.org/abs/1802.08218 2
20. He, K., Zhang, X., Ren, S., Sun, J.: Deep residual learning for image recognition. In: 2016 IEEE Conference on Computer Vision and Pattern Recognition (CVPR). pp. 770–778 (2016). https://doi.org/10.1109/CVPR.2016.90 1, 2
21. Hendrycks, D., Gimpel, K.: A baseline for detecting misclassified and out-of-distribution examples in neural networks (2018), https://arxiv.org/abs/1610.02136 3
22. Howard, J., Ruder, S.: Universal language model fine-tuning for text classification (2018), https://arxiv.org/abs/1801.06146 4
23. Hu, Y., Stretcu, O., Lu, C.T., Viswanathan, K., Hata, K., Luo, E., Krishna, R., Fuxman, A.: Visual program distillation: Distilling tools and programmatic reasoning into vision-language models (2024), https://arxiv.org/abs/2312.03052 11
24. Hudson, D.A., Manning, C.D.: Gqa: A new dataset for real-world visual reasoning and compositional question answering (2019), https://arxiv.org/abs/1902.09506 2
25. Jiang, Y., Natarajan, V., Chen, X., Rohrbach, M., Batra, D., Parikh, D.: Pythia v0.1: the winning entry to the vqa challenge 2018 (2018), https://arxiv.org/abs/1807.09956 11
26. Johnson, J., Hariharan, B., van der Maaten, L., Fei-Fei, L., Zitnick, C.L., Girshick, R.: Clevr: A diagnostic dataset for compositional language and elementary visual reasoning (2016), https://arxiv.org/abs/1612.06890 2
27. Johnson, J., Hariharan, B., van der Maaten, L., Hoffman, J., Fei-Fei, L., Zitnick, C.L., Girshick, R.: Inferring and executing programs for visual reasoning (2017), https://arxiv.org/abs/1705.03633 4
28. Kamath, A., Clark, C., Gupta, T., Kolve, E., Hoiem, D., Kembhavi, A.: Webly supervised concept expansion for general purpose vision models (2022), https://arxiv.org/abs/2202.02317 11
29. Kembhavi, A., Seo, M., Schwenk, D., Choi, J., Farhadi, A., Hajishirzi, H.: Are you smarter than a sixth grader? textbook question answering for multimodal machine comprehension. In: 2017 IEEE Conference on Computer Vision and Pattern Recognition (CVPR). pp. 5376–5384 (2017). https://doi.org/10.1109/CVPR.2017.571 2
30. Khan, Z., BG, V.K., Schulter, S., Chandraker, M., Fu, Y.: Exploring question decomposition for zero-shot vqa (2023), https://arxiv.org/abs/2310.17050 4
31. Kim, T., Cho, Y., Shin, H., Jo, Y., Shin, D.: Generalizing visual question answering from synthetic to human-written questions via a chain of qa with a large language model (2024), https://arxiv.org/abs/2401.06400 4
32. Kim, W., Son, B., Kim, I.: Vilt: Vision-and-language transformer without convolution or region supervision (2021), https://arxiv.org/abs/2102.03334 10
33. Krizhevsky, A., Sutskever, I., Hinton, G.E.: Imagenet classification with deep convolutional neural networks. In: Pereira, F., Burges, C., Bottou, L., Weinberger, K. (eds.) Advances in Neural Information Processing Systems. vol. 25. Curran Associates, Inc. (2012), https://proceedings.neurips.cc/paper_files/paper/2012/file/c399862d3b9d6b76c8436e924a68c45b-Paper.pdf 1
34. Lan, Y., Li, X., Liu, X., Li, Y., Qin, W., Qian, W.: Improving zero-shot visual question answering via large language models with reasoning question prompts (2023), https://arxiv.org/abs/2311.09050 4
35. Le-Quynh, M.D., Nguyen, A.T., Quang-Hoang, A.T., Dinh, V.H., Nguyen, T.H., Ngo, H.B., An, M.H.: Enhancing video retrieval with robust clip-based multimodal system. In: Proceedings of the 12th International Symposium on Information and Communication Technology. pp. 972–979 (2023) 2
36. Lee, K., Lee, H., Lee, K., Shin, J.: Training confidence-calibrated classifiers for detecting out-of-distribution samples (2018), https://arxiv.org/abs/1711.09325 3
37. Li, J., Li, D., Savarese, S., Hoi, S.: Blip-2: Bootstrapping language-image pre-training with frozen image encoders and large language models (2023), https://arxiv.org/abs/2301.12597 10
38. Lin, T.Y., Dollár, P., Girshick, R., He, K., Hariharan, B., Belongie, S.: Feature pyramid networks for object detection (2017), https://arxiv.org/abs/1612.03144 4
39. Liu, H., Li, C., Wu, Q., Lee, Y.J.: Visual instruction tuning (2023), https://arxiv.org/abs/2304.08485 10
40. LMSYS ORG: Vicuna: An open-source chatbot impressing gpt-4 with 90% chatgpt quality (2023), https://lmsys.org/blog/2023-03-30-vicuna/ 10
41. Lu, J., Batra, D., Parikh, D., Lee, S.: Vilbert: Pretraining task-agnostic visiolinguistic representations for vision-and-language tasks (2019), https://arxiv.org/abs/1908.02265 2, 11
42. Marino, K., Chen, X., Parikh, D., Gupta, A., Rohrbach, M.: Krisp: Integrating implicit and symbolic knowledge for open-domain knowledge-based vqa (2020), https://arxiv.org/abs/2012.11014 11
43. Małkiński, M., Mańdziuk, J.: Deep learning methods for abstract visual reasoning: A survey on raven's progressive matrices (2022), https://arxiv.org/abs/2201.12382 4
44. Mokady, R., Hertz, A., Bermano, A.H.: Clipcap: Clip prefix for image captioning (2021), https://arxiv.org/abs/2111.09734 11
45. Nguyen, T.Q., Tran, H.L., Tran, T.K., Phan-Nguyen, H.P., Nguyen, T.H.: Fa-yolov9: Improved yolov9 based on feature attention block. In: 2024 International Conference on Multimedia Analysis and Pattern Recognition (MAPR). pp. 1–6 (2024). https://doi.org/10.1109/MAPR63514.2024.10661057 2

46. Nguyen, T.H., Nguyen-Huu, H.L., Le, T.D., Tran, H.L., Le-Tran, Q.K., Ngo, H.B., An, M.H., Dinh, Q.V.: Multimodal fusion in newsimages 2023: Evaluating translators, keyphrase extraction, and clip pre-training. In: MediaEval (2023) 2
47. Radford, A., Kim, J.W., Hallacy, C., Ramesh, A., Goh, G., Agarwal, S., Sastry, G., Askell, A., Mishkin, P., Clark, J., Krueger, G., Sutskever, I.: Learning transferable visual models from natural language supervision (2021), https://arxiv.org/abs/2103.00020 2, 10
48. Ray, A., Sikka, K., Divakaran, A., Lee, S., Burachas, G.: Sunny and dark outside?! improving answer consistency in vqa through entailed question generation (2019), https://arxiv.org/abs/1909.04696 3
49. Ren, S., He, K., Girshick, R., Sun, J.: Faster r-cnn: Towards real-time object detection with region proposal networks (2016), https://arxiv.org/abs/1506.01497 2
50. Santoro, A., Raposo, D., Barrett, D.G.T., Malinowski, M., Pascanu, R., Battaglia, P., Lillicrap, T.: A simple neural network module for relational reasoning (2017), https://arxiv.org/abs/1706.01427 4
51. Schwartz, R., Dodge, J., Smith, N.A., Etzioni, O.: Green ai (2019), https://arxiv.org/abs/1907.10597 3
52. Schwenk, D., Khandelwal, A., Clark, C., Marino, K., Mottaghi, R.: A-okvqa: A benchmark for visual question answering using world knowledge (2022), https://arxiv.org/abs/2206.01718 9
53. Shah, M., Chen, X., Rohrbach, M., Parikh, D.: Cycle-consistency for robust visual question answering (2019), https://arxiv.org/abs/1902.05660 3
54. Shrestha, R., Kafle, K., Kanan, C.: Answer them all! toward universal visual question answering models (2019), https://arxiv.org/abs/1903.00366 4
55. Suhr, A., Zhou, S., Zhang, A., Zhang, I., Bai, H., Artzi, Y.: A corpus for reasoning about natural language grounded in photographs (2019), https://arxiv.org/abs/1811.00491 4
56. Sun, Q., Fang, Y., Wu, L., Wang, X., Cao, Y.: Eva-clip: Improved training techniques for clip at scale (2023), https://arxiv.org/abs/2303.15389 10
57. Szegedy, C., Liu, W., Jia, Y., Sermanet, P., Reed, S., Anguelov, D., Erhan, D., Vanhoucke, V., Rabinovich, A.: Going deeper with convolutions (2014), https://arxiv.org/abs/1409.4842 4
58. Tan, H., Bansal, M.: Lxmert: Learning cross-modality encoder representations from transformers (2019), https://arxiv.org/abs/1908.07490 11
59. Team, G., Mesnard, T., Hardin, C., Dadashi, R., Bhupatiraju, S., Pathak, S., Sifre, L., Rivière, M., Kale, M.S., Love, J., Tafti, P., Hussenot, L., Sessa, P.G., Chowdhery, A., Roberts, A., Barua, A., Botev, A., Castro-Ros, A., Slone, A., Héliou, A., Tacchetti, A., Bulanova, A., Paterson, A., Tsai, B., Shahriari, B., Lan, C.L., Choquette-Choo, C.A., Crepy, C., Cer, D., Ippolito, D., Reid, D., Buchatskaya, E., Ni, E., Noland, E., Yan, G., Tucker, G., Muraru, G.C., Rozhdestvenskiy, G., Michalewski, H., Tenney, I., Grishchenko, I., Austin, J., Keeling, J., Labanowski, J., Lespiau, J.B., Stanway, J., Brennan, J., Chen, J., Ferret, J., Chiu, J., Mao-Jones, J., Lee, K., Yu, K., Millican, K., Sjoesund, L.L., Lee, L., Dixon, L., Reid, M., Mikuła, M., Wirth, M., Sharman, M., Chinaev, N., Thain, N., Bachem, O., Chang, O., Wahltinez, O., Bailey, P., Michel, P., Yotov, P., Chaabouni, R., Comanescu, R., Jana, R., Anil, R., McIlroy, R., Liu, R., Mullins, R., Smith, S.L., Borgeaud, S., Girgin, S., Douglas, S., Pandya, S., Shakeri, S., De, S., Klimenko, T., Hennigan, T., Feinberg, V., Stokowiec, W., hui Chen, Y., Ahmed, Z., Gong, Z., Warkentin, T., Peran, L., Giang, M., Farabet, C., Vinyals, O., Dean, J., Kavukcuoglu, K.,

Hassabis, D., Ghahramani, Z., Eck, D., Barral, J., Pereira, F., Collins, E., Joulin, A., Fiedel, N., Senter, E., Andreev, A., Kenealy, K.: Gemma: Open models based on gemini research and technology (2024), https://arxiv.org/abs/2403.08295 7, 10

60. Tiong, A.M.H., Li, J., Li, B., Savarese, S., Hoi, S.C.H.: Plug-and-play vqa: Zero-shot vqa by conjoining large pretrained models with zero training (2023), https://arxiv.org/abs/2210.08773 4
61. Touvron, H., Lavril, T., Izacard, G., Martinet, X., Lachaux, M.A., Lacroix, T., Rozière, B., Goyal, N., Hambro, E., Azhar, F., Rodriguez, A., Joulin, A., Grave, E., Lample, G.: Llama: Open and efficient foundation language models (2023), https://arxiv.org/abs/2302.13971 10
62. Tu, H., Cui, C., Wang, Z., Zhou, Y., Zhao, B., Han, J., Zhou, W., Yao, H., Xie, C.: How many unicorns are in this image? a safety evaluation benchmark for vision llms (2023), https://arxiv.org/abs/2311.16101 9
63. Unni, S.J., Moraffah, R., Liu, H.: Vqa-gen: A visual question answering benchmark for domain generalization (2023), https://arxiv.org/abs/2311.00807 4
64. Vaswani, A., Shazeer, N., Parmar, N., Uszkoreit, J., Jones, L., Gomez, A.N., Kaiser, L., Polosukhin, I.: Attention is all you need (2023), https://arxiv.org/abs/1706.03762 1
65. Wang, P., Wu, Q., Shen, C., van den Hengel, A., Dick, A.: Fvqa: Fact-based visual question answering (2017), https://arxiv.org/abs/1606.05433 4
66. Wei, J., Bosma, M., Zhao, V.Y., Guu, K., Yu, A.W., Lester, B., Du, N., Dai, A.M., Le, Q.V.: Finetuned language models are zero-shot learners (2022), https://arxiv.org/abs/2109.01652 10
67. Wu, Q., Teney, D., Wang, P., Shen, C., Dick, A., van den Hengel, A.: Visual question answering: A survey of methods and datasets (2016), https://arxiv.org/abs/1607.05910 2
68. Xie, S., Girshick, R., Dollár, P., Tu, Z., He, K.: Aggregated residual transformations for deep neural networks. In: 2017 IEEE Conference on Computer Vision and Pattern Recognition (CVPR). pp. 5987–5995 (2017). https://doi.org/10.1109/CVPR.2017.634 4
69. Ye, Q., Xu, H., Xu, G., Ye, J., Yan, M., Zhou, Y., Wang, J., Hu, A., Shi, P., Shi, Y., Li, C., Xu, Y., Chen, H., Tian, J., Qian, Q., Zhang, J., Huang, F., Zhou, J.: mplug-owl: Modularization empowers large language models with multimodality (2024), https://arxiv.org/abs/2304.14178 10
70. Ye, Q., Xu, H., Ye, J., Yan, M., Hu, A., Liu, H., Qian, Q., Zhang, J., Huang, F., Zhou, J.: mplug-owl2: Revolutionizing multi-modal large language model with modality collaboration (2023), https://arxiv.org/abs/2311.04257 10
71. Yi, K., Gan, C., Li, Y., Kohli, P., Wu, J., Torralba, A., Tenenbaum, J.B.: Clevrer: Collision events for video representation and reasoning (2020), https://arxiv.org/abs/1910.01442 4
72. Yi, K., Wu, J., Gan, C., Torralba, A., Kohli, P., Tenenbaum, J.B.: Neural-symbolic vqa: Disentangling reasoning from vision and language understanding (2019), https://arxiv.org/abs/1810.02338 4
73. Yu, Z., Ouyang, X., Shao, Z., Wang, M., Yu, J.: Prophet: Prompting large language models with complementary answer heuristics for knowledge-based visual question answering (2023), https://arxiv.org/abs/2303.01903 11
74. Zagoruyko, S., Komodakis, N.: Wide residual networks (2017), https://arxiv.org/abs/1605.07146 4

75. Zakari, R.Y., Owusu, J.W., Wang, H., Qin, K., Lawal, Z.K., Dong, Y.: Vqa and visual reasoning: An overview of recent datasets, methods and challenges (2022), https://arxiv.org/abs/2212.13296 4
76. Zhang, P., Dong, X., Wang, B., Cao, Y., Xu, C., Ouyang, L., Zhao, Z., Duan, H., Zhang, S., Ding, S., Zhang, W., Yan, H., Zhang, X., Li, W., Li, J., Chen, K., He, C., Zhang, X., Qiao, Y., Lin, D., Wang, J.: Internlm-xcomposer: A vision-language large model for advanced text-image comprehension and composition (2023), https://arxiv.org/abs/2309.15112 10
77. Zhang, R., Han, J., Liu, C., Gao, P., Zhou, A., Hu, X., Yan, S., Lu, P., Li, H., Qiao, Y.: Llama-adapter: Efficient fine-tuning of language models with zero-init attention (2023), https://arxiv.org/abs/2303.16199 10
78. Zhao, H., Shi, J., Qi, X., Wang, X., Jia, J.: Pyramid scene parsing network. In: 2017 IEEE Conference on Computer Vision and Pattern Recognition (CVPR). pp. 6230–6239 (2017). https://doi.org/10.1109/CVPR.2017.660 4
79. Zhong, Z., Zheng, L., Kang, G., Li, S., Yang, Y.: Random erasing data augmentation. Proceedings of the AAAI Conference on Artificial Intelligence **34** (08 2017). https://doi.org/10.1609/aaai.v34i07.7000 4
80. Zhou, Z.H.: Ensemble Methods: Foundations and Algorithms, vol. 14 (06 2012). https://doi.org/10.1201/b12207 4
81. Zhu, D., Chen, J., Shen, X., Li, X., Elhoseiny, M.: Minigpt-4: Enhancing vision-language understanding with advanced large language models (2023), https://arxiv.org/abs/2304.10592 10